\documentclass{article}

\usepackage[final,nonatbib]{neurips_2023}

\usepackage[utf8]{inputenc} 
\usepackage[T1]{fontenc}    
\usepackage{url}            
\usepackage{booktabs}       
\usepackage{amsfonts}       
\usepackage{nicefrac}       
\usepackage{microtype}      
\usepackage{xcolor}     
\usepackage{hyperref}       
\usepackage{multibib}

\usepackage{todonotes}
\usepackage{enumitem}

\usepackage{tikz}
\usepackage{pgfplots}
\usepackage{amsmath}
\usetikzlibrary{plotmarks}
\usepgfplotslibrary{fillbetween}
\usetikzlibrary{arrows, decorations.markings, decorations}
\usetikzlibrary{arrows,arrows.meta,patterns,positioning,shapes}
\usetikzlibrary{plotmarks}
\usepgfplotslibrary{fillbetween}
\usetikzlibrary{arrows, decorations.markings, decorations}
\usepackage{mathtools}
\usepackage{acro}
\usepackage{booktabs}
\usepackage{pgfplots}
\usepgfplotslibrary{groupplots,dateplot}
\usepackage[all]{nowidow}
\usepackage{soul}
\usepackage{tabularx,multirow}
\usepackage{xspace}
\usepackage{tikz}
\usepackage{bm}
\usetikzlibrary{arrows,arrows.meta,patterns,positioning,shapes}
\usepackage{soul}
\usepackage{enumitem}
\usepackage[ruled,vlined]{algorithm2e}
\usepackage{makecell}
\usepackage{siunitx}
\usepackage{multirow}
\usepackage{adjustbox}
\usepackage{csvsimple}
\usepackage[capitalise]{cleveref}

\newcommand{\bftab}{\fontseries{b}\selectfont}
\pgfplotsset{
    compat=1.15,
}

\definecolor{tabblue}{HTML}{5075b2}
\definecolor{tabred}{HTML}{bf575a}
\definecolor{tabgreen}{HTML}{71b582}
\definecolor{tabyellow}{HTML}{cdba77}
\DeclareSIUnit{\pp}{\textup{p.p.}}

\usepackage{acro}
\DeclareAcronym{FL}{short=FL,long=federated learning,short-indefinite=an}
\DeclareAcronym{IOT}{short=IoT,long=internet of things,short-indefinite=an}
\DeclareAcronym{NN}{short=NN,long=neural network,short-indefinite=an}
\DeclareAcronym{CNN}{short=CNN, long=convolutional neural network}
\DeclareAcronym{FEDAVG}{short=FedAvg, long=Federated Averaging}
\DeclareAcronym{IID}{short=iid, long=independent and identically distributed}
\DeclareAcronym{ML}{short=ML, long=machine learning}
\DeclareAcronym{KD}{short=KD, long=knowledge distillation}
\DeclareAcronym{SGD}{short=SGD, long=stochastic gradient descent}
\DeclareAcronym{NAS}{short=NAS, long=neural architecture search}
\DeclareAcronym{FD}{short=FD, long=Federated Dropout}
\DeclareAcronym{FLOP}{short=FLOP, long=Floating Point Operation}
\newcommand{\etal}{\emph{et al.}\xspace}

\DeclareAcronym{SLT}{short=SLT, long=Successive Layer Training}

\title{Aggregating Capacity in FL through Successive Layer Training for Computationally-Constrained Devices}

\author{%
  Kilian Pfeiffer \\
  Karlsruhe Institute of Technology\\
  Karlsruhe, Germany\\
  \texttt{kilian.pfeiffer@kit.edu} \\
  \And
  Ramin Khalili \\
  Huawei Research Center Munich \\
  Munich, Germany\\
  \texttt{ramin.khalili@huawei.com} \\
  \And
  J{\"o}rg Henkel  \\
  Karlsruhe Institute of Technology\\
  Karlsruhe, Germany\\
  \texttt{henkel@kit.edu} \\
}

\begin{document}

\maketitle

\begin{abstract}
\Ac{FL} is usually performed on resource-constrained edge devices, e.g., with limited memory for the computation. If the required memory to train a model exceeds this limit, the device will be excluded from the training.
This can lead to a lower accuracy as valuable data and computation resources are excluded from training, also causing bias and unfairness. 
The \ac{FL} training process should be adjusted to such constraints.
The state-of-the-art techniques propose training subsets of the \ac{FL} model at constrained devices, reducing their resource requirements for training. However, these techniques largely limit the co-adaptation among parameters of the model and are highly inefficient, as we show: it is actually better to train a smaller (less accurate) model by the system where all the devices can train the model end-to-end than applying such techniques. 
We propose a new method that enables successive freezing and training of the parameters of the \ac{FL} model at devices, reducing the training's resource requirements at the devices while still allowing enough co-adaptation between parameters. 
We show through extensive experimental evaluation that our technique greatly improves the accuracy of the trained model (by~$\SI{52.4}{\pp}$) compared with the state of the art, efficiently aggregating the computation capacity available on distributed devices.
\end{abstract}

\acresetall

\section{Introduction}
\Ac{FL} has achieved impressive results in many domains and is proposed for several use cases, such as healthcare, transportation, and robotics~\cite{chen2020fedhealth, yuan2020federated,ciftler2020federated,posner2021federated, liu2019lifelong, liu2019federated}. As data in \ac{FL} is not processed centrally but usually on \emph{resource-constrained} edge devices, training \ac{ML} models impose a large computational burden on these devices~\cite{samie2016computation}. Additionally, \ac{FL} requires communication, specifically exchanging \ac{ML} model parameters from the devices to a centralized entity for aggregation. Extensive research has been done to lower the communication overhead required for \ac{FL}, e.g., on the use of quantization in the communication~\cite{mills2019communication, caldas2018expanding} or sketched updates~\cite{rothchild2020fetchsgd}. Similarly, techniques such as partial updates~\cite{li2020federated}, asynchronous aggregation~\cite{chen2019asynchronous, xie2020asynchronous}, and tier-based aggregation~\cite{chai2020fedat, Chai2020} have been proposed to lower and account for varying computational throughput. While constrained computation throughput and communication capabilities can slow down \ac{FL} convergence, high memory requirements for training that are imposed on devices can exclude devices completely from the \ac{FL} system. This is, for example, the case in \emph{Google GBoard}~\cite{yang2018applied}, where devices that do not have 2GB of memory for training are removed. Excluding devices from training lowers the reachable accuracy, as fewer devices participate in the training, also resulting  in bias and unfairness~\cite{maeng2022towards}. 

\begin{figure}[h]
    \centering
    \includegraphics[page=1]{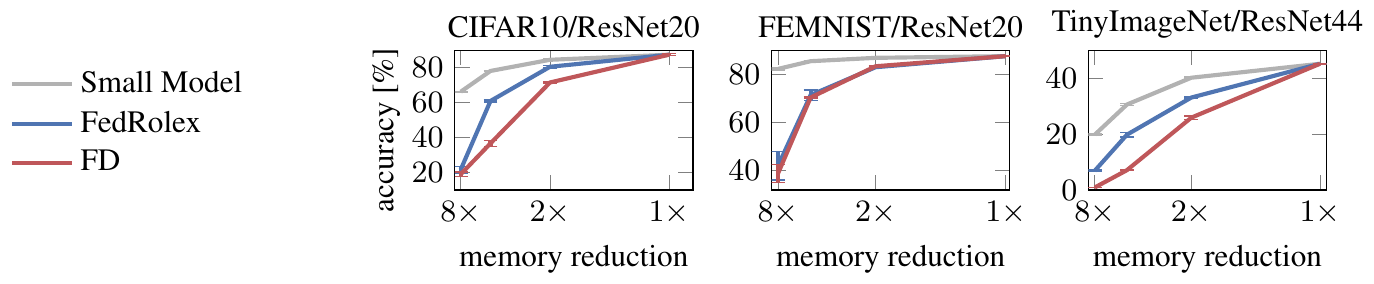} 
    \caption{Accuracy of FedRolex~\cite{alam2022fedrolex} and \ac{FD}~\cite{caldas2018expanding} compared to a small model using different \ac{NN} topologies and datasets after $2500/1000$~\ac{FL} rounds. In the case of $1\times$, both techniques are equivalent to vanilla \ac{FEDAVG}. Hyperparameters of the experiments are given in \cref{sec:experiments}.}
    \label{fig:motivational_example}
\end{figure}

Several techniques have been proposed to tackle these constraints, where the main idea is to train a lower complexity \emph{submodel} on the devices and embed the trained submodel into the full higher-capacity server model. A submodel is typically created by \textit{scaling the width} of the \ac{NN}, e.g., using a subset of convolutional filters per \ac{NN} layer. There exist several variations of the technique~\cite{caldas2018expanding, diao2020heterofl, horvath2021fjord, alam2022fedrolex}. In particular,  Caldas~\etal~\cite{caldas2018expanding} propose \textit{\ac{FD}}, which randomly, per round and per device, selects \ac{NN} filters that are trained. Alam~\etal~\cite{alam2022fedrolex} propose \emph{FedRolex}, a sliding window approach, where all devices train the same submodel, and in each \ac{FL} round, the used filter indices are shifted by one. While both these techniques allow training within given memory constraints, our results show (\cref{fig:motivational_example}) that they perform worse than a straightforward baseline, i.e., using a \emph{smaller} \ac{NN} model that can be trained by all devices end-to-end.
We evaluate CIFAR10, FEMNIST, and TinyImageNet in \iac{FL} setting using ResNet and scale the width of the \ac{NN} down s.t. we achieve a~$2-8\times$ reduction in training memory. We observe that training the same small model at all devices outperforms FedRolex and \ac{FD} w.r.t to the final accuracy and convergence speed (we expect similar results for other subset-derived techniques), especially when enforcing a~$4\times$ and~$8\times$ memory reduction (as also evaluated in~\cite{alam2022fedrolex}).
\\
Our results indicate that applying these techniques is rather harmful. This is as a large part of filters/parameters has to be dropped during each round of training at each device, extremely limiting the co-adaptation between parameters. Hence, the gradients for the subset of parameters that are trained on devices are calculated without considering the error of the parameters that reside on the server (more details in~\cref{appdx:co_adaption}). Motivated by these observations, we propose a new technique that enables successive freezing and training of the parameters of the \ac{FL} model at devices, reducing the training’s resource requirements at the devices while allowing a higher co-adaptation between parameters. Instead of switching between subsets of the model on an \ac{FL}-round basis, we train the same parameters for several rounds and successively switch to a larger model. To obey the same memory constraints as in~\cite{caldas2018expanding, alam2022fedrolex}, we train early layers using the full width while utilizing a scaled-down \ac{NN} head. We then freeze the early layers and expand the head layers' width. By freezing early layers, no activation has to be kept in memory, hence, we lower the memory footprint. But still, the error of these frozen parameters is included in the calculation of the gradient of the new subset of parameters. We apply this technique successively till all parameters of the model are trained. 

In summary, we make the following novel contributions:
\begin{itemize}[noitemsep, nolistsep]
    \item We empirically show that employing current state-of-the-art techniques, \ac{FD}~\cite{caldas2018expanding} and FedRolex~\cite{alam2022fedrolex}, for memory-constrained systems can actually hurt the performance.
    \item We propose a novel training scheme called \textit{\ac{SLT}}\footnote{The source code of \ac{SLT} is available at \url{https://github.com/k1l1/SLT}.}, which addresses the shortcomings of previous techniques by successively adding more parameters to the training, successively freezing layers, and reusing a scaled-down \ac{NN} head. 
    \item Our evaluation of common \ac{NN} topologies, such as ResNet and DenseNet, shows that \ac{SLT} reaches significantly higher accuracies in \ac{IID} and non-\ac{IID} CIFAR, FEMNIST, and TinyImageNet training compared to the state of the art. Also, \ac{SLT} provides a much faster convergence, reducing the communication overhead to reach a certain level of accuracy by over~$10\times$ compared with \ac{FD} and FedRolex. The same behavior can be observed w.r.t. \acp{FLOP}, where \ac{SLT} requires~$10\times$ fewer operations to reach a certain level of accuracy compared with \ac{FD} and FedRolex.
    \item We study the performance of \ac{SLT} in heterogeneous settings. We show that devices with different memory constraints can make a meaningful contribution to the global model, significantly outperforming the state-of-the-art techniques.    
\end{itemize}

\section{Methodology}
\label{sec:methodology}
\subsection{Problem Statement and Setup}
We consider a synchronous cross-device \ac{FL} setting, where we have one \emph{server} and a set of \textit{devices}~$c \in \mathcal{C}$ as participants. There is a given \ac{ML} model topology~$F$ on the \ac{FL} server that is trained in a distributed manner for~$R$ rounds. Our goal is to maximize the accuracy of the model. Similar to \ac{FD} and FedRolex, we assume that a fixed number of devices~$|\mathcal{C}^{(r)}|$ out of~$\mathcal{C}$ participate in a round~$r\leq R$. All devices are constrained in memory, and thus their training must not exceed this given memory constraint~$m_{\text{constraint}}$. In other terms, we assume that no participating device can train the server \ac{NN} model end-to-end.
\SetKwComment{Comment}{// }{}
\SetKwProg{kwDevice}{DeviceTraining}{:}{}
\SetKwProg{kwServer}{Server}{:}{}
\SetKwProg{kwConfig}{ConfigurationSelection}{:}{}
\SetKwInput{Input}{Requires}
\begin{algorithm}[t]
\SetAlgoLined
\DontPrintSemicolon
\caption{Successive Layer Training: $w$ and $W$ label the set of all layers' parameters.}\label{alg:algorithm}
\Input{Number of rounds~$R$, devices~$\mathcal C$, number of devices per round~$|\mathcal{C}^{(r)}|$, configurations~$S_n \ n \in [1,\ldots,N]$, that satisfy constraint $m$, init. parameters~$W^{(1)}$}
\kwServer{}{
\ForEach{\normalfont{round} $r=1,2,\ldots, R$}{
$\mathcal{C}^{(r)}$ $\gets$ select $|\mathcal{C}^{(r)}|$ random devices out of $\mathcal C$\;
$w^{(r)}, S^{(r)}$ $\gets$ ConfigurationSelection($W^{(r)}, r$)\;
\ForEach{device $c \in \mathcal C^{(r)}$ in parallel}{
$w^{(r)}, S^{(r)}$ receive from server\;
$w^{(r,c)} \gets$ \normalfont{DeviceTraining}$(w^{(r)}, S^{(r)})$\;
upload $w^{(r,c)}_j$ to server $\forall j \in \{j: K_F < j\leq K \}$\;
}
$w_j^{(r+1)} \gets \frac{1}{|\mathcal{C}^{(r)}|}\sum \limits_{c \in \mathcal{C}^{(r)}} w_j^{(r,c)}$ $\quad$\texttt{//averaging of trained layers}\;
$W^{(r+1)} \gets w^{(r+1)}$ $\quad$\texttt{//layers get embedded into server model}\;
}
}
\kwConfig{($W$, $r$)}{
$S \gets \text{LookupTable}(r, W)$\;
$w \gets W$ $\quad$$\quad$$\quad$$\quad$\texttt{//scaling down \ac{NN} head based on configuration}\;
Return $w, S$\;
}
\kwDevice{$(w, S)$}{
freeze $w_j$ for $j\in \{0 < j \leq K_F\}$ according to $S$\;
\ForEach{\normalfont{local mini-batch} $b$}{
    $w \gets w - \eta \nabla l(w, b)$
}
Return $w$\;
}
\end{algorithm}
\subsection{Successive Layer Training}
\begin{figure}
    \centering
    \includegraphics[page=1]{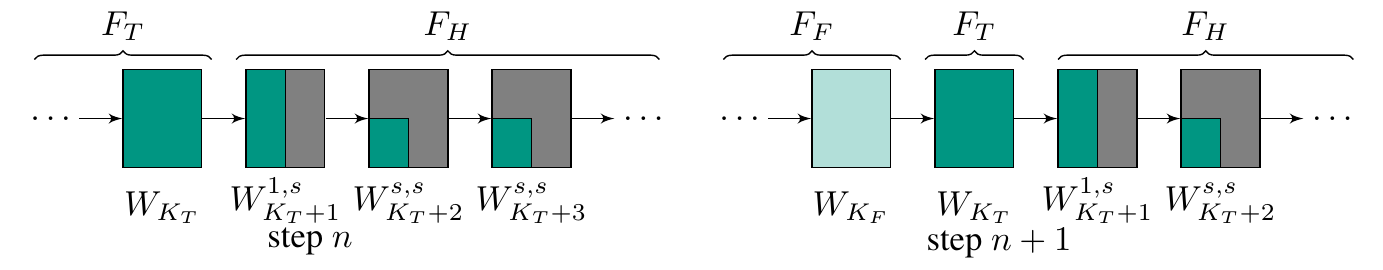} 
    \caption{Visualization of \ac{SLT}. With each step,~$K_F$ and $K_T$ are shifted by~$1$. $W_1,\ldots,W_{K_F}$ denote the parameters that remain frozen during training,~$W_{K_T}$ denotes parameters of layer~$K_T$ that are fully trained, while~$W_{K_T+1}^{1,s},\ldots,W_{K}^{s,s}$ denote the parameters of the scaled-down head using~$s$.}
    \label{fig:training_configuration}
\end{figure}

The following describes our methodology of \textit{\acl{SLT}} for \acp{CNN}. Firstly, we rewrite~$F$ such that it is the consecutive operations of~$K$ \emph{layers},
where each layer is defined as~$f_k$, $k\in[1,\cdots, K]$. Each layer~$f_k$ has associated server parameters~$W_k$. We label a convolution, followed by batch normalization and an activation function, a layer. Similar to~\cite{caldas2018expanding, alam2022fedrolex}, we define a \textit{subset}~$w_k$ of the layer parameters (server)~$W_k$ that is \textit{scaled} down using~$s$ as
\begin{equation}
    w_k = W_k^{s,s} \quad w_k \in \mathbb R^{\lfloor sP_k \rfloor \times \lfloor sM_k \rfloor} \quad W_k \in \mathbb R^{P_k \times M_k},
\end{equation}
where~$P_k$ labels the layer's input dimension, $M_k$ labels the output dimension of the fully-sized server parameters, and~$s \in (0,1]$ is a scaling factor (we omit the filter kernel
dimensions for brevity). 
To obey the memory constraint on the participating devices,
we split the \ac{NN} into three consecutive parts. The first part of the \ac{NN} contains layers that are already trained and remain frozen on the devices. The second part contains layers that are being fully trained on the devices. The last part represents the \ac{NN}'s \textit{head}. To train the remaining layers within the memory budget, the head's parameters are scaled down. Throughout the \ac{FL} training, we successively switch the \textit{training configuration}, s.t., the part of the \ac{NN} that is being fully trained~($s=1$) moves from the first layer to the last layer. Thereby, successively, the remaining parameters from the scaled-down head are added. At the same time, we successively freeze more layers, starting with the first layer, to stay within the memory budget. We visualize the switching from one training configuration to the next in~\cref{fig:training_configuration}. The parts of the \ac{NN} that are frozen, trained, and represent the head are labeled $F_F$, $F_T$, and $F_H$. The resulting model that is trained on the devices can be described as  $F = F_F \circ F_T \circ F_H$: 
\begin{itemize}[noitemsep, nolistsep]
    \item The first part~$F_F$ labels the part of the \ac{NN} that is frozen and where the server parameters~$W_1,\ldots,W_{K_F}$ do not get updated by the devices. Freezing the first part of the \ac{NN} reduces the memory overhead during training, as in the forward pass, activations do not have to be stored for frozen layers. The frozen part of the \ac{NN} is defined as~$F_F := \bigcirc_{k \in \{k: 0 < k \leq K_F\}} f_k$, where layers~$1,\ldots,K_F$ remain frozen.
    \item The second part~$F_T$ labels the part of the \ac{NN} that is being fully trained by the devices. The parameters~$W_{K_F+1},\ldots, W_{K_T}$ get updated during training. This part is defined as~$F_T := \bigcirc_{k \in \{k: K_F < k \leq K_T\}} f_k$, s.t. layers $K_F + 1,\ldots,K_T$ are fully trained.
    \item The last part~$F_H$ describes the \ac{NN}'s head that is scaled down using~$s$, s.t.~$F_H := \bigcirc_{k \in \{k: K_T < k \leq K\}} f_k$, where the scaled-down layers~$K_T +1, \ldots, K$ are trained. The first layer of~$F_{H}$ scales down the parameters to the width of the head s.t.~$w_{K_T+1} = W_{K_T+1}^{1,s}$, where $W_{K_T+1}^{1,s} \in \mathbb R^{P_{K_T+1} \times \lfloor sM_{K_T+1} \rfloor}$. All consecutive layers are scaled down using~$s$, s.t.~$ w_{K_T+j} = W_{K_T+j}^{s,s} \forall j \in [2, \ldots , K - K_T]$.
\end{itemize}
We define a set~$S$ as a training \emph{configuration} that obeys the given memory constraint, s.t.~$S = \{K_F, K_T, s\}$ fully describes the mapping of layers $f_k, k \in [1,\ldots, K]$ into $F_F$, $F_T$, $F_H$, and the head's scale factor~$s$. Each~$S$ has a respective memory footprint during training.~$m = \text{memory}(S)$ denotes the maximum memory that is utilized during training for a given configuration~$S$. The maximum memory of a configuration can be determined by measurements or by calculating the size of weights, gradients, and activations that have to be kept in memory (see \cref{subsec:memory_training}).

\subsection{Configuration Selection}
\begin{figure}
    \centering
    \includegraphics[page=1]{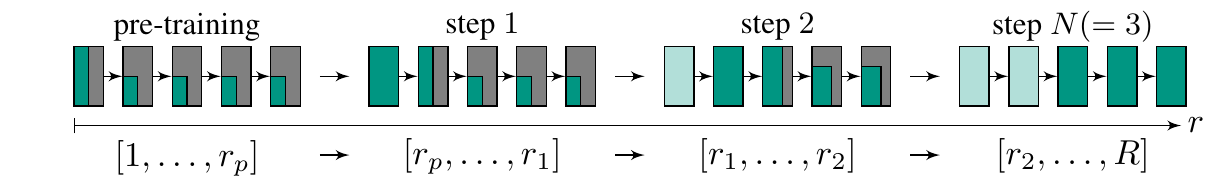} 
    \caption{Visualization of the \ac{SLT} training scheme with an exemplary $5$-layer \ac{NN}. The model is first pre-trained for~$r_p$ rounds. Following that, the model is trained for $r_1-r_p$, $r_2-r_1$, and~$R-r_2$ rounds in configuration $1$, $2$, and~$N(=3)$, respectively.}
    \label{fig:training_configuration2}
\end{figure}

For each selected~$S$, we aim to fully utilize the available memory. We define $n$ as a \emph{configuration step} in $ n \in [1,\ldots,N]$, where we add parameters to the training (i.e., \textit{fill up} the remaining parameters of a head's layer). These steps are distributed over the total training rounds~$R$ (~\cref{fig:training_configuration2}). We set $K_T = K_F+1$, s.t. in each configuration step exactly one layer gets fully trained (filled up).
We start with $K_F=K_T=0$ (consequently $F = F_H$) to pre-train the head for a certain number of rounds. After pre-training, we increase~$K_F$ by one and apply~$K_T = K_F+1$ (the first configuration has no frozen layers, i.e., $F = F_T \circ F_H$) and continue training\footnote{We discuss in the evaluation section how to decide the number of rounds a certain configuration should be trained before switching to the next configuration.}. We switch to the successive configuration by increasing $K_F$ by one. Hence, for the training configuration at step $n$, we have $K_F=n-1$ and $K_T = n$, with $s_n$ selected as follows:   
\begin{align}
\label{eq:constraint_definition}
    \text{max} \ s_n, \ \text{s.t.} \quad \text{memory}(S_n) \leq m_{\text{constraint}} \wedge s_n \leq s_j \quad \forall j \in [n+1,\ldots,N],
\end{align}
\Cref{eq:constraint_definition} ensures that each configuration obeys the constraint $m_{\text{constraint}}$. The second constraint in \cref{eq:constraint_definition} enforces that $s_n$ can only grow with increasing $n$ to ensure that parameters of the head are only added throughout the training but not removed. We provide a justification for maximizing~$s$ instead of~$F_T$ by performing an ablation study in~\cref{appdx:ablation_study1}.
The configuration selection is performed \emph{offline}. 
Lastly, we define the last step $N$ where a given memory constraint ($\text{memory}(S_N)$) allows for~$s=1$. If this step is reached, we train with $S_N$ for all remaining rounds since the memory budget allows to fully train all remaining parameters at once. We provide a visualization of the training process in~\cref{fig:training_configuration2} and outline \ac{SLT} in~\cref{alg:algorithm}.  

\section{Experimental Evaluation}
\label{sec:experiments}

\subsection{Experimental Setting and Hyperparamters}
\label{subsec:hyperparameters}
We evaluate \ac{SLT} in an \ac{FL} setting using PyTorch~\cite{PyTorch}, where we distribute a share from the datasets CIFAR10, CIFAR100~\cite{krizhevsky2009learning}, FEMNIST from the Leaf~\cite{caldas1812leaf} benchmark, and TinyImageNet~\cite{le2015tiny} to each device~$c \in \mathcal{C}$, s.t. each device~$c$ has a local dataset~$\mathcal{D}_c$ of the same size. In each round~$r$, a subset of devices~$\mathcal{C}^{(r)}$ is selected. We train with the optimizer \ac{SGD} with momentum of~$0.9$, an initial learning rate of~$\eta = 0.1$, and apply cosine annealing to~$\eta = 0.01$ and a weight decay of~$1.0\times 10^{-5}$. We evaluate the vision models ResNet20, ResNet44~\cite{he2015deep}, and DenseNet40~\cite{densenet}. For each experiment, we report the average accuracy and standard deviation of~$3$ independent seeds after~$R$ rounds of \ac{FL} training. For CIFAR10, CIFAR100, and TinyImageNet, we evaluate a scenario with~$|\mathcal{C}| = 100$ devices, where each round~$|\mathcal{C}^{(r)}|=10$ devices are actively participating. For FEMNIST, we evaluate with~$|\mathcal{C}| = 3550$ and $|\mathcal{C}^{(r)}|=35$. We train for~$R=2500$ rounds for CIFAR10, CIFAR100, and TinyImageNet, and~$R=1000$ for FEMNIST. In each round~$r$, each participating device iterates once over its local dataset. We apply standard image augmentation techniques like random cropping and horizontal and vertical flips to all datasets (horizontal and vertical flips are omitted for FEMNIST). An input resolution of~$3\times32\times32$ is used for CIFAR and FEMNIST (up-scaled from $28\times28$) and~$3\times64\times64$ for TinyImageNet. We use batch size~$32$ and perform~$363$ experiments in total, with an average run-time of~$\SI{6}{\hour}$ on an NVIDIA Tesla V100.

\textbf{Comparison with the state of the art:}
We compare \ac{SLT} against several baselines. We introduce~$I^{(r)}_k$ as the set of indices of the output dimension of a layer~$k \in [1,\ldots,K]$ where~$r \in [1,\ldots, R]$ denotes the rounds. Consequently, for full-sized \acp{NN}~$|I^{(r)}_k| = M_k$. The subset for training is scaled down by building a dense matrix using the indices from~$I^{(r)}_k$, s.t. the scaled-down parameters are~$w_k \in \mathbb R ^{\lfloor sP_k \rfloor \times \lfloor sM_k \rfloor}$. The consecutive layer's input dimension indices are equal to the last layer's output indices. The first layer's input dimension is not scaled to feed all color channels into the \ac{NN}.

\textbf{Small model}: Devices train a submodel where all filters per layer are scaled down by~$s$, s.t. all devices can train the submodel. The remaining filters are trained end-to-end throughout the rounds. The same submodel is used for evaluation. The indices of the output dimension of a layer~$k$ are selected using~$I_k^{(r)}=I_k=\{i: \ 0 \leq i < \lfloor sM_k \rfloor\}$.

\textbf{FedRolex~\cite{alam2022fedrolex}:} FedRolex creates a submodel by scaling the numbers of filters using~$s$. Each device trains the same continuous block of indices. The filter indices trained on the devices are shifted in a rolling window fashion every round. The server averages the trained block and evaluates using the full server model~($s=1$). The indices~$I_k^{(r)}$ are selected with $\hat r = r \mod M_k$ using
\begin{align}
\label{eq:fedrolex}
    I_k^{(r)} = \begin{cases} 
    \{ \hat r, \hat r+1, \ldots, r + \lfloor s M_k \rfloor -1 \} & \text{if } \hat r + \lfloor s M_k \rfloor \leq M_k \\
    \{ \hat r, \hat r+1, \ldots, M_k -1 \} \cup \{0,\ldots, \hat r + \lfloor s M_k  \rfloor - 1 - M_k \} & \text{otherwise}
    \end{cases}.
\end{align}
\textbf{\ac{FD}~\cite{caldas2018expanding}:} FD creates a submodel by scaling down the number of filters using~$s$. The indices of the filters are randomly sampled per device per round on the server. Hence, the indices~$I_k^{(r,c)}$ of a device~$c$ of round~$r$ is a round-based per-device random selection of~$\lfloor sM_k \rfloor$ indices out of all~$M_k$ indices. The server aggregates the device-specific submodels after training and evaluates the full model ($s=1$). 

\subsection{Memory Footprint during Training}
\label{subsec:memory_training}
The high memory requirements during training can be split into three groups: Firstly, the weights of the \ac{NN} have to be stored in memory. This is required both for the forward pass and the backward pass. Secondly, for the calculated gradients in the backward pass, the activation maps of the respective layers have to be kept in memory. Lastly, the calculated gradients have to be stored in memory. In state-of-the-art CNNs, the size of the activation map makes up for most of the memory requirements, while the size of the weights only plays a minor role. For ResNet44 and DenseNet40, we measure that activations make up for $\sim99\%$ of the required memory for training, while gradients and parameters account for the remaining $1\%$. Consequently, the required memory linearly reduces with $s$ for \ac{FD} and FedRolex, as the number of layer's output channels~$\lfloor sM_k \rfloor$ determines the activation map's size. Similarly, for \ac{SLT}, we measure the maximum amount of memory that is required during training by counting the size of the activation maps, as well as the loaded weights and gradients in training. For frozen layers, it is only required to load the parameters in memory, while no activation maps and gradients have to be stored. For the fully trained layer~$K_T$, it is required to store the layer's full parameters $w_{K_T}$, as well as the full-size activation map and gradients in memory. For all other layers (\ac{NN} head), memory scales linearly with $s$. We provide implementation details in \cref{appdx:pytorch}.

We evaluate memory constraints that are given by scaling down~$s$ in FedRolex and \ac{FD} by~$s_{\text{FD/FedRolex}} \in [0.125, 0.25, 0.5, 1.0]$ for experiments with ResNet and $[0.33, 0.66, 1.0]$ for DenseNet\footnote{For DenseNet, \ac{SLT} only enables a reduction of $3\times$, as in DenseNet, specific layers have a significantly larger sized feature map than others, which limits our technique's effectiveness w.r.t memory reduction.}.  
In \ac{SLT}, for a given~$s_{\text{FD/FedRolex}}$, we adjust~$s_n$ for each step $n$ in the following way
\begin{align}
    \text{max} \ s_n, \ \text{s.t.} \quad \text{memory}(S_n) \leq \text{memory}(s_{\text{FD/FedRolex}}) \wedge s_n \leq s_j \quad \forall j \in [n+1,\ldots,N],
\end{align}
to ensure that our technique obeys the same constraint as the baselines. If $s_{\text{FD/FedRolex}} = 1.0$, all algorithms coincide with vanilla \ac{FEDAVG} using the full server model. 

We distribute the required steps~$N$ over the total rounds~$R$, s.t. all parameters receive sufficient training. Specifically, we distribute the rounds based on the share of parameters that are added to the training within a configuration~$n$. We calculate the number of all trained parameters $Q$ by using $K_F$,$K_T$, and $s$ s.t.
\begin{equation}
	Q(K_F, K_T, s) = \bigg(\smashoperator[r]{\sum_{k \in \{k:K_F < k \leq K_T\}}} P_k M_k \bigg) + P_{K_T+1}\lfloor sM_{K_T+1}\rfloor \quad +\quad \smashoperator{\sum_{k \in \{k:K_T+1 < k \leq K\}}}\lfloor sP_k\rfloor \lfloor sM_k \rfloor,
\end{equation}
and use $Q$ to calculate the share of rounds $R_n$ for a step $n$. The share of rounds for pretraining is calculated using $R_\text{pretraining} = R \frac{Q(0,0,s)}{Q(0,0,1)}$. For step 1, $R_1= R\frac{Q(0,1,s)}{Q(0,0,1)} - R_\text{pretraining}$. For all steps $n > 1$, we calculate the rounds using
\begin{equation}
 R_n = R \frac{Q(n-1,n,s_n) - Q(n-2,n-1,s_{n-1})}{Q(0,0,1)}.
\end{equation}
Lastly, the switching point for pretraining is $r_\text{pretraining} = R_\text{pretraining}$ and $r_n = R \frac{Q(n-1,n,s_n)}{Q(0,0,1)}$ for all steps~$n$.

Preliminary experiments have shown that this mapping scheme outperforms other techniques, like an equal distribution of rounds to all configurations, and enables \ac{SLT} to converge as fast as a small model while reaching a significantly higher final accuracy (we provide further results in~\cref{appdx:mapping}). The mapping of steps~$N$ to rounds~$R$ does not rely on private data (or any data) and can be stored in a look-up table prior to the training. A visualization of \ac{SLT} is given in \cref{fig:training_configuration2}. We provide the number of steps $N$ for different \ac{NN} architectures and constraints in \cref{appdx:mapping}.

\subsection{Experimental Results}
\textbf{\Ac{IID} results:} For the \ac{IID} case, results are given in \cref{tab:acc}. We observe that \ac{SLT} reaches significantly higher accuracy for ResNet20 and CIFAR10 for all evaluated constraints, outperforming a small model baseline by up to~$\SI{7.8}{\pp}$ and state of the art by~$\SI{52.4}{\pp}$. The results with FEMNIST show that a small model baseline already provides sufficient capacity for the dataset since only a few percentage points separate $s_{\text{FD/FedRolex}} = 0.125$ from the full model (i.e., when $s_{\text{FD/FedRolex}} = 1$). Hence, \ac{SLT} can only provide a minor benefit over a small model baseline. The contrary can be observed for CIFAR100 and TinyImageNet, where using a small model~($s_{\text{FD/FedRolex}} = 0.125$) loses up to $\SI{25.4}{\pp}$ to the full model. Additionally, it can be observed that for low memory constraints, FD and FedRolex fail to learn a useful representation at all. \ac{SLT} improves upon a small model by up to $\SI{13.7}{\pp}$ and up to $\SI{26.6}{\pp}$ compared to state of the art.
\begin{table*}
\centering
\caption{Results for \ac{IID} experiments with ResNet and DenseNet using CIFAR10, FEMNIST, CIFAR100, and TinyImageNet. Accuracy in~$\%$ after $R$ rounds of training is given.}
\label{tab:acc}
\color{black}
\begin{adjustbox}{width=\columnwidth,center}
\small{
\begin{tabular}{l c c c c c c c c}
\toprule
Setting &
\multicolumn{4}{c}{\textbf{ResNet20/CIFAR10}}&\multicolumn{4}{c}{\textbf{ResNet20/FEMNIST}}\\ 

\cmidrule(l{1pt}r{5pt}){2-5} \cmidrule(l{1pt}r{5pt}){6-9}

$s_{\text{FD/FedRolex}}$ & 0.125 & 0.25 & 0.5 & 1.0 &
0.125 & 0.25 & 0.5 & 1.0\\
\midrule

\ac{SLT} (ours) &\textbf{74.1}$\pm$\bftab{0.8} &  \bftab{83.4}$\pm$\bftab{0.2} & 
\bftab{85.2}$\pm$\bftab{0.6} & 
\multirow{ 4}{*}{87.5$\pm$0.6} & 

\bftab{84.4}$\pm$\bftab{0.3} & 
\bftab{85.8}$\pm$\bftab{0.1} & 
\bftab{86.9}$\pm$\bftab{0.0}&
\multirow{ 4}{*}{87.6$\pm$0.0}\\

Small model&
66.3$\pm$0.3 &
78.2$\pm$0.4 &
84.6$\pm$0.4 & &

82.3$\pm$0.4 &
85.5$\pm$0.1 &
86.9$\pm$0.0 &\\

FedRolex~\cite{alam2022fedrolex} &
21.7$\pm$1.9 &
61.0$\pm$0.5 &
80.6$\pm$0.6 & &

42.1$\pm$6.0 &
71.4$\pm$2.1 &
83.0$\pm$0.1 &\\

\ac{FD}~\cite{caldas2018expanding} &
19.0$\pm$1.4 &
36.7$\pm$1.8 &
71.6$\pm$0.4 &  &

38.9$\pm$3.7 &
70.4$\pm$0.2 &
83.4$\pm$0.3 &\\

\midrule
Setting & &
\multicolumn{3}{c}{\textbf{DenseNet40/CIFAR100}}&\multicolumn{4}{c}{\textbf{ResNet44/TinyImageNet}}\\ 

\cmidrule(l{1pt}r{5pt}){3-5} \cmidrule(l{1pt}r{5pt}){6-9}

$s_{\text{FD/FedRolex}}$ &  & 0.33 & 0.66 & 1.0 &
0.125 & 0.25 & 0.5 & 1.0\\
\midrule

\ac{SLT} (ours) & &  \bftab{51.1}$\pm$\bftab{0.4} & 
53.3$\pm$0.6 & 
\multirow{ 4}{*}{60.2$\pm$0.5} & 

\bftab{33.5}$\pm$\bftab{0.1} & 
\bftab{40.3}$\pm$\bftab{0.5} & 
\bftab{42.3}$\pm$\bftab{0.2} &
\multirow{ 4}{*}{45.2$\pm$0.1}\\

Small model& &
43.9$\pm$1.5 &
\bftab{55.9}$\pm$\bftab{0.1} & &

19.8$\pm$0.3 &
30.6$\pm$0.3 &
40.2$\pm$0.3 &\\

FedRolex~\cite{alam2022fedrolex} & &
22.2$\pm$0.3 &
46.7$\pm$0.1 & &

\phantom{0}6.9$\pm$0.2 &
19.8$\pm$0.8 &
33.1$\pm$0.2 &\\

\ac{FD}~\cite{caldas2018expanding} & &
13.5$\pm$0.5 &
41.9$\pm$1.5 &  &

\phantom{0}0.9$\pm$0.0 &
\phantom{0}7.1$\pm$0.1 &
25.9$\pm$0.6 &\\

\bottomrule
\end{tabular}
}
\end{adjustbox}
\end{table*}

\textbf{Non-\ac{IID} results:}
Typically, data in \ac{FL} is not distributed in an \ac{IID} fashion but rather non-\ac{IID}. We repeat the experiments shown in~\cref{tab:acc} but distribute the data on the devices in a non-\ac{IID} fashion. Similar to~\cite{hsu2019measuring}, we apply a Dirichlet distribution, where the rate of non-\ac{IID}-ness can be varied using~$\alpha$. For all experiments, we set~$\alpha=0.1$. We observe from \Cref{tab:acc_noniid} that the small model baselines in the case of CIFAR10 and FEMNIST lose accuracy compared to the full model. Hence, the gain of \ac{SLT} compared to a small model baseline increases. The results for CIFAR100 and TinyImageNet show a proportional drop in accuracy for all algorithms. However,  \ac{SLT} still outperforms other techniques by a large margin. 
Note that we could apply common non-\ac{IID} mitigation techniques like FedProx~\cite{li2020federated} on top of \ac{SLT} to further limit the drop in accuracy.
\begin{table*}
\centering
\caption{Results for non-\ac{IID} experiments with ResNet and DenseNet using CIFAR10, FEMNIST, CIFAR100, and TinyImageNet. Accuracy in~$\%$ after $R$ rounds of training is given.}
\label{tab:acc_noniid}
\color{black}
\begin{adjustbox}{width=\columnwidth,center}
\small{
\begin{tabular}{l c c c c c c c c}
\toprule
Setting &
\multicolumn{4}{c}{\textbf{ResNet20/CIFAR10}}&\multicolumn{4}{c}{\textbf{ResNet20/FEMNIST}}\\ 

\cmidrule(l{1pt}r{5pt}){2-5} \cmidrule(l{1pt}r{5pt}){6-9}

$s_{\text{FD/FedRolex}}$ & 0.125 & 0.25 & 0.5 & 1.0 &
0.125 & 0.25 & 0.5 & 1.0\\
\midrule

\ac{SLT} (ours) &\textbf{52.4}$\pm$\bftab{0.9} &  \bftab{69.6}$\pm$\bftab{0.6} & 
\bftab{75.5}$\pm$\bftab{1.3} & 
\multirow{ 4}{*}{80.5$\pm$1.3} & 

\bftab{81.2}$\pm$\bftab{1.6} & 
\bftab{83.0}$\pm$\bftab{2.0} & 
\bftab{83.8}$\pm$\bftab{1.9} &
\multirow{ 4}{*}{84.0$\pm$1.9}\\

Small model&
44.7$\pm$1.2 &
63.1$\pm$0.7 &
73.6$\pm$0.6 & &

79.6$\pm$0.8 &
82.9$\pm$1.1 &
83.3$\pm$2.4 &\\

FedRolex~\cite{alam2022fedrolex} &
15.0$\pm$3.7 &
29.8$\pm$1.7 &
48.3$\pm$2.9 & &

39.4$\pm$2.0 &
59.3$\pm$2.1 &
78.5$\pm$0.5 &\\

\ac{FD}~\cite{caldas2018expanding} &
11.3$\pm$0.9 &
10.7$\pm$0.6 &
34.9$\pm$5.7 &  &

15.9$\pm$8.2 &
51.0$\pm$1.2 &
79.7$\pm$1.1 &\\

\midrule
Setting & &
\multicolumn{3}{c}{\textbf{DenseNet40/CIFAR100}}&\multicolumn{4}{c}{\textbf{ResNet44/TinyImageNet}}\\ 

\cmidrule(l{1pt}r{5pt}){3-5} \cmidrule(l{1pt}r{5pt}){6-9}

$s_{\text{FD/FedRolex}}$ &  & 0.33 & 0.66 & 1.0 &
0.125 & 0.25 & 0.5 & 1.0\\
\midrule

\ac{SLT} (ours) &  &  \bftab{45.9}$\pm$\bftab{1.4} & 
48.4$\pm$0.5 & 
\multirow{ 4}{*}{55.8$\pm$0.5} & 

\bftab{28.5}$\pm$\bftab{1.2} & 
\bftab{35.1}$\pm$\bftab{1.1} & 
\bftab{36.1}$\pm$\bftab{0.2} &
\multirow{ 4}{*}{39.0$\pm$0.8}\\

Small model& &
40.5$\pm$1.2 &
\bftab{51.8}$\pm$\bftab{0.2} & &

16.9$\pm$0.2 &
25.3$\pm$0.5 &
34.2$\pm$0.4 &\\

FedRolex~\cite{alam2022fedrolex} & &
20.0$\pm$0.3 &
42.9$\pm$0.4 & &

\phantom{0}1.5$\pm$0.5 &
12.7$\pm$1.3 &
26.1$\pm$0.5 &\\

\ac{FD}~\cite{caldas2018expanding} & &
7.6$\pm$0.1 &
36.9$\pm$1.0 &  &

\phantom{0}0.4$\pm$0.1 &
\phantom{0}0.6$\pm$0.0 &
20.0$\pm$1.3 &\\

\bottomrule
\end{tabular}
}
\end{adjustbox}
\end{table*}

For additional experimental results, we refer the readers to \cref{appdx:other_results}. 

\begin{figure}
    \centering
    \includegraphics[page=1]{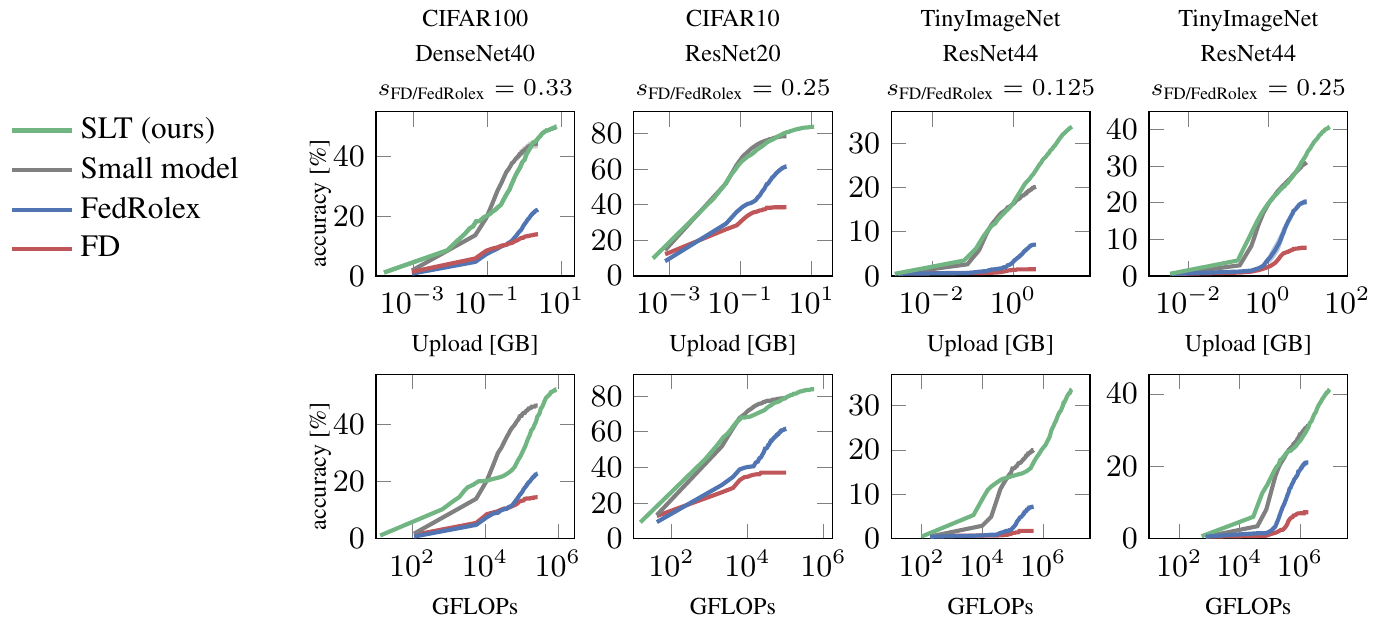} 
    \caption{Maximum reached accuracy in \% over data upload and performed \acp{FLOP} for CIFAR10, CIFAR100, and TinyImageNet using DenseNet40, ResNet20, and ResNet44 in an \ac{IID} training case.}
    \label{fig:communication}
\end{figure}

\textbf{Communication, computation, and convergence speed:}
We evaluate our technique w.r.t. the communication overhead of the distributed training and the number of computations devices have to perform (\acp{FLOP}). Specifically, we evaluate the gain in accuracy over required transmitted data and performed \acp{FLOP}. We show the results in~\cref{fig:communication}.
We observe that our technique converges fast, similarly to a small model, while reaching higher final accuracy. Compared to FD and FedRolex, our technique requires significantly less communication to reach the same level of accuracy. Similar behavior can be observed w.r.t. \acp{FLOP}.
\subsection{Heterogeneous Memory Constraints}
Memory constraints in devices can be heterogeneous. We evaluate \ac{SLT} in such scenarios and compare it against the start of the art. We evaluate with different resource levels and split the available constraint levels equally upon the devices, i.e., when constraints of~$\bm s_{\text{FD/FedRolex}} = [0.125, 0.25]$ are given,~$50\%$ of the devices train with~$s_{\text{FD/FedRolex}}=0.125$ while the remaining~$50\%$ use~$s_{\text{FD/FedRolex}}=0.25$. \Ac{SLT} supports heterogeneous constraints through the following mechanism: Firstly, devices with the highest constraint perform training as done in the homogeneous case, outlined in \cref{alg:algorithm} using the head's scale factor as described in \cref{eq:constraint_definition}. Devices that are less constrained use the same scale factor $s_n$ per configuration to ensure that all devices train the same number of parameters within a layer. To utilize the remaining available memory, less constrained devices freeze fewer layers, therefore, train more layers at full width. For a given $K_T$ and $s_n$ of a configuration $S_n$, the remaining memory of less constrained devices is utilized by minimizing $K_F$, s.t.
\begin{align}
\label{eq:constraint_heterogeneous}
    \text{min} \ K_F, \ \text{s.t.} \quad \text{memory}(S_n) \leq \text{memory}(s_{\text{FD/FedRolex}}).
\end{align}
In addition to \ac{FD} and FedRolex, we evaluate HeteroFL~\cite{diao2020heterofl} and FjORD~\cite{horvath2021fjord}. Both require that some devices are capable of training the full \ac{NN} end-to-end, otherwise, some parameters do not receive any updates. In cases where no device can train the server model end-to-end, we reduce the size of the server model such that at least one participating device can fully train the model.

\textbf{Small model}: All devices train a small model regardless of their constraints. Scale factor~$s$ is set to the minimum a participating device supports.

\textbf{FedRolex}~\cite{alam2022fedrolex}: Similar to the homogeneous case, FedRolex uses a rolling window (\cref{eq:fedrolex}). In heterogeneous cases, devices use a constraint-specific scale~$s_e$, to adjust the number of filters~$\lfloor s_eM_k \rfloor$.

\textbf{\ac{FD}}~\cite{caldas2018expanding}: Although Caldas~\etal\cite{caldas2018expanding} do not specifically evaluate heterogeneous devices, heterogeneity can be supported straightforwardly by using constraint-specific $s_e$ for scaling down the \ac{NN}. This extension of \ac{FD} is also evaluated in FedRolex and FjORD.

\textbf{HeteroFL}~\cite{diao2020heterofl}: In HeteroFL, devices use the same subset throughout the training. To support heterogeneity, devices use a resource-specific scaling factor $s_e$, s.t. for each $s_e$ the indices are selected using~$I_k^{(r,e)}=I_k^{(e)}=\{i \ | \ 0 \leq i < \lfloor s_eM_k \rfloor\}$.

\textbf{FjORD}~\cite{horvath2021fjord}: FjORD uses the same indices as HeteroFL, but each device switches between constraint-specific subsets that satisfy the device constraints within a local epoch on a mini-batch level.

\textbf{Heterogeneity results}: We repeat the experimental setup as presented in \cref{tab:acc_noniid} for TinyImageNet and CIFAR100, but enforce varying device constraints in the experiments (see \cref{tab:acc_heterogeneous}). We observe that \ac{SLT} outperforms others in all evaluated scenarios. FedRolex can improve upon a small model in some settings, but this is not the case with \ac{FD}. For FjORD and HeteroFL, we observe that both outperform the small model baseline. Yet, in some cases, both HeteroFL and FjORD have a lower accuracy when utilizing more constraint levels. For HeteroFL, it can be observed that using ResNet44 with 4 constraint levels reaches a lower accuracy than with 3 levels (despite the fact that all devices have higher average resources $\mathbb E[\bm s_{\text{FD/FedRolex}}]$ of $\approx 0.47$ in the case of $[0.125, 0.25, 0.5, 1.0]$ instead of $\approx0.29$ in the case of $[0.125, 0.25, 0.5]$). The same can be observed for FjORD with DenseNet40. As both techniques, in principle, use the same subset mechanism as FedRolex and \ac{FD}, we think that both suffer from supporting more constraint levels that cause less co-adaptation between \ac{NN} filters.

\begin{table*}
\centering
\caption{\ac{FL} with heterogeneous constraints. Accuracy in $\%$ after $R$ rounds of training is given.}
\label{tab:acc_heterogeneous}
\color{black}
\begin{adjustbox}{width=\columnwidth,center}
\small{
\begin{tabular}{l c c c c c}
\toprule
Setting &
\multicolumn{2}{c}{\textbf{DenseNet40/CIFAR100}} & \multicolumn{3}{c}{\textbf{ResNet44/TinyImageNet}} \\ 

\cmidrule(l{1pt}r{5pt}){2-3} 
\cmidrule(l{1pt}r{5pt}){4-6}

$\bm s_{\text{FD/FedRolex}}$ & [0.33, 0.66] & [0.33, 0.66, 1.0] & [0.125, 0.25] & [0.125, 0.25, 0.5] & [0.125, 0.25, 0.5, 1.0]\\
\midrule

\ac{SLT} (ours) &\textbf{46.4}$\pm$\bftab{2.0}&  \bftab{49.3}$\pm$\bftab{1.8}& 
\bftab{30.3}$\pm$\bftab{1.2} & 
\bftab{33.0}$\pm$\bftab{0.5} & \bftab{35.9}$\pm$\bftab{0.4}\\

Small model& 40.5$\pm$1.2 & 40.5$\pm$1.2 &
16.9$\pm$0.2 &16.9$\pm$0.2 &16.9$\pm$0.2\\

FedRolex~\cite{alam2022fedrolex} & 33.2$\pm$0.4 &43.9$\pm$1.3 &
\phantom{0}5.4$\pm$0.2 &13.8$\pm$1.3&23.6$\pm$0.7\\

\ac{FD}~\cite{caldas2018expanding} & 21.2$\pm$0.6 &38.1$\pm$0.4 &
\phantom{0}0.5$\pm$0.1 &\phantom{0}0.6$\pm$0.1 &20.6$\pm$1.7\\

HeteroFL~\cite{diao2020heterofl} & 42.2$\pm$1.3 &42.8$\pm$0.5 &
20.7$\pm$0.7 &24.1$\pm$0.2 &23.3$\pm$0.5\\

FjORD~\cite{horvath2021fjord} & 38.7$\pm$0.4 &36.9$\pm$0.4 &
22.4$\pm$1.0 &25.3$\pm$0.3 &27.5$\pm$0.8\\

\bottomrule
\end{tabular}
}
\end{adjustbox}
\end{table*}

\section{Related Work}
We cover related work that studies similar problems or employs similar techniques. 

\textbf{Resource constraints in \ac{FL}}: Most works on resource-constrained \ac{FL} target communication. Specifically, the use of quantization and compression in communication~\cite{caldas2018expanding, mills2019communication} and sketched updates~\cite{rothchild2020fetchsgd} have been proposed to lower the communication burden. Chen~\etal~\cite{chen2021communication} propose \textit{adaptive parameter freezing} as they discover that parameters stabilize during training and do not have to be transferred to the server. Another branch of work focuses on reducing communication, computation, and memory requirements by employing only a subset of the full \ac{NN} on devices. Caldas~\etal~\cite{caldas2018expanding} introduce \emph{\ac{FD}}, a mechanism that creates a device-specific subset by randomly selecting a subset of \ac{CNN} filters. Diao~\etal~\cite{diao2020heterofl} introduce HeteroFL, which allows for heterogeneous constraints by employing fixed subsets of different sizes to the \ac{NN} and aggregating them on the server. Horvath~\etal~\cite{horvath2021fjord} introduce a similar technique (FjORD), with the main difference that in FjORD, each device trains every available subset within its capabilities. Rapp~\etal~\cite{rapp2021distreal} propose DISTREAL, a technique that uses varying subsets on a mini-batch level such that devices finish their update on time despite having intra-round changing resources. FedRolex~\cite{alam2022fedrolex} supports heterogeneity similar to FjORD and HeteroFL but allows for server \ac{NN} models that are outside of the capabilities of all devices. This is enabled by not training a fixed subset of the \ac{NN} parameters but by training a rolling window of all parameters that is shifted on a round basis. Beyond subsets, the use of low-rank factorization~\cite{yao2021fedhm,mei2022resource} has been proposed to train \ac{NN} models. Lastly, Qui~\etal~\cite{qiu2022zerofl} propose sparse convolutions to lower the resource requirements for training but require special hardware for sparse computations to realize the gains.

\textbf{Layer-wise model training:}
Layer-wise model training has been proposed in centralized training of \ac{CNN}s as an alternative to training with end-to-end backpropagation of the error. Hettinger~\etal~\cite{hettinger2017forward} introduced a technique that adds \ac{CNN} layers one at a time during training using auxiliary heads for classification while freezing early layers. Similar techniques have also been employed for unsupervised learning, where representations are trained with contrastive techniques without requiring end-to-end gradient propagation~\cite{lowe2019putting, xiong2020loco}. Recently, the concept of progressive model growth has also been proposed for \ac{FL}: Wang~\etal~\cite{wang2022progfed} propose \emph{ProgFed}, where they discover that by progressively adding \ac{CNN} layers to the \ac{NN} while using an auxiliary head, the \ac{FL} training converges faster and required less communication to reach the same accuracy as an end-to-end baseline. Similarly, Kundu~\etal~\cite{kundu2022fednet2net} propose a technique that grows the model depending on the complexity of the data to reach a high accuracy if the \ac{NN} capacity is not sufficient for the problem. \textit{Importantly, both techniques only focus on increasing the convergence speed. Hence, they consider communication and computation overhead but not the problem of constrained memory on edge devices, nor do they support heterogeneous devices. In both techniques, eventually, all devices have to train the full-size \ac{NN} and, consequently, have to have the memory resources available for that.}

\textbf{Memory-efficient training}: Several techniques have been proposed to train an \ac{ML} model in a memory-efficient way. Kirisame~\etal~\cite{kirisamedynamic} present Dynamic Tensor Rematerialization that allows recomputing activation maps on the fly. Similarly, encoding and compression schemes~\cite{jain2018gist, Georgiadis_2019_CVPR} have been proposed to lower the size of the activation maps during training. Techniques like that trade memory for computation,  and some lower the accuracy by using approximation or lossy compression. \textit{Importantly, these techniques are orthogonal to \ac{SLT}, \ac{FD}, FedRolex, HeteroFL, and FjORD.}

\section{Conclusion}
We proposed \ac{SLT} that is able to reduce the memory requirements for training on devices, efficiently aggregating computation capacity and learning from all available data. Through extensive evaluation, we show that gains in final accuracy as well as the faster convergence speed (compared with state of the art) are robust throughout different datasets, data distribution, and \ac{NN} topologies.

\textbf{Limitations:} We observe that \ac{SLT} is most effective if the used \ac{NN} architecture is deep (i.e., has many layers), as the cost of \emph{filling up} a single layer becomes less significant. Also, \ac{SLT} is less effective if the size of the activation map is strongly unevenly distributed throughout the layers (DenseNet), as it has to adapt to the layer with the highest memory requirements when filled up. Besides, we applied \ac{SLT} to CNN topologies only. Finally, we mainly focused on memory as a \textit{hard constraint}~\cite{pfeiffer2023federated} for training. We show that communication and \ac{FLOP} efficiency are significantly higher than in the state of the art, but we did not consider per-round communication or \ac{FLOP} constraints. For future work, we want to extend our study to other topologies, such as transformers, and employ \ac{NAS} techniques to find \ac{NN} configurations that reach the highest accuracy when trained in \ac{FL} with \ac{SLT} in heterogeneous environments.

\textbf{Broader impact:} Our solution could help reduce biases in \ac{FL} systems, improving fairness. For instance, by including users with low-end smartphones in the learning process, it provides these users (who perhaps cannot afford high-end devices) with better experiences as the model is going to be trained over their data, too. It could also reduce the cost of deployment of distributed IoT systems (e.g., sensor networks), as they can be implemented with low-cost devices (or a mixture of low and high-cost devices), enabling, e.g., deployment of larger and more fine-grained monitoring systems. On the negative side, distributing learning over low-end devices that are not particularly designed for training tasks can increase the overall energy consumption of the system. This is an important issue that should be studied in more detail.

\begin{ack}
This work was partially funded by the “Helmholtz Pilot Program for Core Informatics (kikit)” at Karlsruhe Institute of Technology. The authors acknowledge support by the state of Baden-Württemberg through bwHPC.
\end{ack}

\bibliographystyle{unsrt}
\bibliography{bib/bibfile}

\begin{thebibliography}{10}

\bibitem{chen2020fedhealth}
Yiqiang Chen, Xin Qin, Jindong Wang, Chaohui Yu, and Wen Gao.
\newblock Fedhealth: A federated transfer learning framework for wearable
  healthcare.
\newblock {\em IEEE Intelligent Systems}, 35(4):83--93, 2020.

\bibitem{yuan2020federated}
Binhang Yuan, Song Ge, and Wenhui Xing.
\newblock A federated learning framework for healthcare iot devices.
\newblock {\em arXiv:2005.05083}, 2020.

\bibitem{ciftler2020federated}
Bekir~Sait Ciftler, Abdullatif Albaseer, Noureddine Lasla, and Mohamed
  Abdallah.
\newblock Federated learning for localization: A privacy-preserving
  crowdsourcing method.
\newblock {\em arXiv:2001.01911}, 2020.

\bibitem{posner2021federated}
Jason Posner, Lewis Tseng, Moayad Aloqaily, and Yaser Jararweh.
\newblock Federated learning in vehicular networks: Opportunities and
  solutions.
\newblock {\em IEEE Network}, 2021.

\bibitem{liu2019lifelong}
Boyi Liu, Lujia Wang, and Ming Liu.
\newblock Lifelong federated reinforcement learning: a learning architecture
  for navigation in cloud robotic systems.
\newblock {\em IEEE Robotics and Automation Letters}, 4(4):4555--4562, 2019.

\bibitem{liu2019federated}
Boyi Liu, Lujia Wang, Ming Liu, and Cheng-Zhong Xu.
\newblock Federated imitation learning: A privacy considered imitation learning
  framework for cloud robotic systems with heterogeneous sensor data.
\newblock {\em arXiv:1909.00895}, 2019.

\bibitem{samie2016computation}
Farzad Samie, Vasileios Tsoutsouras, Lars Bauer, Sotirios Xydis, Dimitrios
  Soudris, and J{\"o}rg Henkel.
\newblock Computation offloading and resource allocation for low-power iot edge
  devices.
\newblock In {\em 2016 IEEE 3rd world forum on internet of things (WF-IoT)},
  pages 7--12. IEEE, 2016.

\bibitem{mills2019communication}
Jed Mills, Jia Hu, and Geyong Min.
\newblock Communication-efficient federated learning for wireless edge
  intelligence in iot.
\newblock {\em IEEE Internet of Things Journal}, 7(7):5986--5994, 2019.

\bibitem{caldas2018expanding}
Sebastian Caldas, Jakub Kone{\v{c}}ny, H~Brendan McMahan, and Ameet Talwalkar.
\newblock Expanding the reach of federated learning by reducing client resource
  requirements.
\newblock {\em arXiv:1812.07210}, 2018.

\bibitem{rothchild2020fetchsgd}
Daniel Rothchild, Ashwinee Panda, Enayat Ullah, Nikita Ivkin, Ion Stoica,
  Vladimir Braverman, Joseph Gonzalez, and Raman Arora.
\newblock Fetchsgd: Communication-efficient federated learning with sketching.
\newblock In {\em International Conference on Machine Learning}, pages
  8253--8265. PMLR, 2020.

\bibitem{li2020federated}
Tian Li, Anit~Kumar Sahu, Manzil Zaheer, Maziar Sanjabi, Ameet Talwalkar, and
  Virginia Smith.
\newblock Federated optimization in heterogeneous networks.
\newblock In {\em Proceedings of Machine Learning and Systems}, volume~2, pages
  429--450, 2020.

\bibitem{chen2019asynchronous}
Y.~{Chen}, Y.~{Ning}, M.~{Slawski}, and H.~{Rangwala}.
\newblock Asynchronous online federated learning for edge devices with non-iid
  data.
\newblock In {\em 2020 IEEE International Conference on Big Data (Big Data)},
  pages 15--24, 2020.

\bibitem{xie2020asynchronous}
Cong Xie, Sanmi Koyejo, and Indranil Gupta.
\newblock Asynchronous federated optimization.
\newblock {\em arXiv:1903.03934}, 2020.

\bibitem{chai2020fedat}
Zheng Chai, Yujing Chen, Liang Zhao, Yue Cheng, and Huzefa Rangwala.
\newblock Fedat: A communication-efficient federated learning method with
  asynchronous tiers under non-iid data.
\newblock {\em arXiv:2010.05958}, 2020.

\bibitem{Chai2020}
Zheng Chai, Ahsan Ali, Syed Zawad, Stacey Truex, Ali Anwar, Nathalie Baracaldo,
  Yi~Zhou, Heiko Ludwig, Feng Yan, and Yue Cheng.
\newblock Tifl: A tier-based federated learning system.
\newblock In {\em Proceedings of the 29th International Symposium on
  High-Performance Parallel and Distributed Computing}, HPDC '20, page
  125–136, New York, NY, USA, 2020. Association for Computing Machinery.

\bibitem{yang2018applied}
Timothy Yang, Galen Andrew, Hubert Eichner, Haicheng Sun, Wei Li, Nicholas
  Kong, Daniel Ramage, and Françoise Beaufays.
\newblock Applied federated learning: Improving google keyboard query
  suggestions.
\newblock {\em arXiv:1812.02903}, 2018.

\bibitem{maeng2022towards}
Kiwan Maeng, Haiyu Lu, Luca Melis, John Nguyen, Mike Rabbat, and Carole-Jean
  Wu.
\newblock Towards fair federated recommendation learning: Characterizing the
  inter-dependence of system and data heterogeneity.
\newblock In {\em Proceedings of the 16th ACM Conference on Recommender
  Systems}, pages 156--167, 2022.

\bibitem{diao2020heterofl}
Enmao Diao, Jie Ding, and Vahid Tarokh.
\newblock Heterofl: Computation and communication efficient federated learning
  for heterogeneous clients.
\newblock In {\em International Conference on Learning Representations}, 2020.

\bibitem{horvath2021fjord}
Samuel Horvath, Stefanos Laskaridis, Mario Almeida, Ilias Leontiadis, Stylianos
  Venieris, and Nicholas Lane.
\newblock Fjord: Fair and accurate federated learning under heterogeneous
  targets with ordered dropout.
\newblock {\em Advances in Neural Information Processing Systems},
  34:12876--12889, 2021.

\bibitem{alam2022fedrolex}
Samiul Alam, Luyang Liu, Ming Yan, and Mi~Zhang.
\newblock Fedrolex: Model-heterogeneous federated learning with rolling
  sub-model extraction.
\newblock In {\em Advances in Neural Information Processing Systems},
  volume~35, pages 29677--29690, 2022.

\bibitem{PyTorch}
Adam Paszke, Sam Gross, Francisco Massa, Adam Lerer, James Bradbury, Gregory
  Chanan, Trevor Killeen, Zeming Lin, Natalia Gimelshein, Luca Antiga, Alban
  Desmaison, Andreas Kopf, Edward Yang, Zachary DeVito, Martin Raison, Alykhan
  Tejani, Sasank Chilamkurthy, Benoit Steiner, Lu~Fang, Junjie Bai, and Soumith
  Chintala.
\newblock Pytorch: An imperative style, high-performance deep learning library.
\newblock In H.~Wallach, H.~Larochelle, A.~Beygelzimer, F.~d\textquotesingle
  Alch\'{e}-Buc, E.~Fox, and R.~Garnett, editors, {\em Advances in Neural
  Information Processing Systems}, volume~32, pages 8024--8035, 2019.

\bibitem{krizhevsky2009learning}
Alex Krizhevsky, Geoffrey Hinton, et~al.
\newblock Learning multiple layers of features from tiny images, 2009.

\bibitem{caldas1812leaf}
Sebastian Caldas, Sai Meher~Karthik Duddu, Peter Wu, Tian Li, Jakub
  Kone{\v{c}}n{\`y}, H~Brendan McMahan, Virginia Smith, and Ameet Talwalkar.
\newblock Leaf: A benchmark for federated settings.
\newblock {\em arXiv:1812.01097}, 2019.

\bibitem{le2015tiny}
Ya~Le and Xuan~S. Yang.
\newblock Tiny imagenet visual recognition challenge, 2015.

\bibitem{he2015deep}
Kaiming He, Xiangyu Zhang, Shaoqing Ren, and Jian Sun.
\newblock Deep residual learning for image recognition.
\newblock In {\em Proceedings of the IEEE conference on computer vision and
  pattern recognition}, pages 770--778, 2016.

\bibitem{densenet}
Gao Huang, Zhuang Liu, Laurens Van Der~Maaten, and Kilian~Q. Weinberger.
\newblock Densely connected convolutional networks.
\newblock In {\em Conference on Computer Vision and Pattern Recognition
  (CVPR)}, pages 2261--2269, 2017.

\bibitem{hsu2019measuring}
Tzu-Ming~Harry Hsu, Hang Qi, and Matthew Brown.
\newblock Measuring the effects of non-identical data distribution for
  federated visual classification.
\newblock {\em arXiv:1909.06335}, 2019.

\bibitem{chen2021communication}
Chen Chen, Hong Xu, Wei Wang, Baochun Li, Bo~Li, Li~Chen, and Gong Zhang.
\newblock Communication-efficient federated learning with adaptive parameter
  freezing.
\newblock In {\em 2021 IEEE 41st International Conference on Distributed
  Computing Systems (ICDCS)}, pages 1--11. IEEE, 2021.

\bibitem{rapp2021distreal}
Martin Rapp, Ramin Khalili, Kilian Pfeiffer, and J{\"o}rg Henkel.
\newblock Distreal: Distributed resource-aware learning in heterogeneous
  systems.
\newblock In {\em Proceedings of the AAAI Conference on Artificial
  Intelligence}, volume~36, pages 8062--8071, 2022.

\bibitem{yao2021fedhm}
Dezhong Yao, Wanning Pan, Yao Wan, Hai Jin, and Lichao Sun.
\newblock Fedhm: Efficient federated learning for heterogeneous models via
  low-rank factorization.
\newblock {\em arXiv:2111.14655}, 2021.

\bibitem{mei2022resource}
Yiqun Mei, Pengfei Guo, Mo~Zhou, and Vishal Patel.
\newblock Resource-adaptive federated learning with all-in-one neural
  composition.
\newblock In {\em Advances in Neural Information Processing Systems},
  volume~35, pages 4270--4284, 2022.

\bibitem{qiu2022zerofl}
Xinchi Qiu, Javier Fernandez-Marques, Pedro~PB Gusmao, Yan Gao, Titouan
  Parcollet, and Nicholas~Donald Lane.
\newblock Zerofl: Efficient on-device training for federated learning with
  local sparsity.
\newblock In {\em International Conference on Learning Representations}, 2022.

\bibitem{hettinger2017forward}
Chris Hettinger, Tanner Christensen, Ben Ehlert, Jeffrey Humpherys, Tyler
  Jarvis, and Sean Wade.
\newblock Forward thinking: Building and training neural networks one layer at
  a time.
\newblock {\em arXiv:1706.02480}, 2017.

\bibitem{lowe2019putting}
Sindy L{\"o}we, Peter O'Connor, and Bastiaan Veeling.
\newblock Putting an end to end-to-end: Gradient-isolated learning of
  representations.
\newblock {\em Advances in Neural Information Processing Systems},
  32:3039--3051, 2019.

\bibitem{xiong2020loco}
Yuwen Xiong, Mengye Ren, and Raquel Urtasun.
\newblock Loco: Local contrastive representation learning.
\newblock {\em Advances in neural information processing systems},
  33:11142--11153, 2020.

\bibitem{wang2022progfed}
Hui-Po Wang, Sebastian Stich, Yang He, and Mario Fritz.
\newblock Progfed: effective, communication, and computation efficient
  federated learning by progressive training.
\newblock In {\em International Conference on Machine Learning}, pages
  23034--23054. PMLR, 2022.

\bibitem{kundu2022fednet2net}
Amit~Kumar Kundu and Joseph Jaja.
\newblock Fednet2net: Saving communication and computations in federated
  learning with model growing.
\newblock In {\em Artificial Neural Networks and Machine Learning--ICANN 2022:
  31st International Conference on Artificial Neural Networks, Bristol, UK,
  September 6--9, 2022, Proceedings; Part IV}, pages 236--247. Springer, 2022.

\bibitem{kirisamedynamic}
Marisa Kirisame, Steven Lyubomirsky, Altan Haan, Jennifer Brennan, Mike He,
  Jared Roesch, Tianqi Chen, and Zachary Tatlock.
\newblock Dynamic tensor rematerialization.
\newblock In {\em International Conference on Learning Representations}, 2021.

\bibitem{jain2018gist}
Animesh Jain, Amar Phanishayee, Jason Mars, Lingjia Tang, and Gennady
  Pekhimenko.
\newblock Gist: Efficient data encoding for deep neural network training.
\newblock In {\em 2018 ACM/IEEE 45th Annual International Symposium on Computer
  Architecture (ISCA)}, pages 776--789. IEEE, 2018.

\bibitem{Georgiadis_2019_CVPR}
Georgios Georgiadis.
\newblock Accelerating convolutional neural networks via activation map
  compression.
\newblock In {\em Proceedings of the IEEE/CVF Conference on Computer Vision and
  Pattern Recognition (CVPR)}, June 2019.

\bibitem{pfeiffer2023federated}
Kilian Pfeiffer, Martin Rapp, Ramin Khalili, and J\"{o}rg Henkel.
\newblock Federated learning for computationally constrained heterogeneous
  devices: A survey.
\newblock {\em ACM Comput. Surv.}, 55(14s), jul 2023.

\bibitem{srivastava2014dropout}
Nitish Srivastava, Geoffrey Hinton, Alex Krizhevsky, Ilya Sutskever, and Ruslan
  Salakhutdinov.
\newblock Dropout: a simple way to prevent neural networks from overfitting.
\newblock {\em The journal of machine learning research}, 15(1):1929--1958,
  2014.

\bibitem{deng2009imagenet}
Jia Deng, Wei Dong, Richard Socher, Li-Jia Li, Kai Li, and Li~Fei-Fei.
\newblock Imagenet: A large-scale hierarchical image database.
\newblock In {\em 2009 IEEE conference on computer vision and pattern
  recognition}, pages 248--255. Ieee, 2009.

\end{thebibliography}

\newpage

\appendix

\section{Co-Adaptation in Subset-based \ac{FL}}
\label{appdx:co_adaption}
\Acl{FD} is originally inspired by regular dropout~\cite{srivastava2014dropout}, a regularization technique that constrains the capacity of a large \ac{NN} model by randomly dropping parameters from the training, thereby limiting co-adaptation among parameters of the model. This is essential to improve the accuracy and reduce the over-fitting of parameters, as shown in various studies. \ac{FD} and FedRolex adopt the dropout technique, removing CNN's filters in a round-based manner. These techniques, however, exercise dropout to its extreme, dropping a large part of filters so that not enough co-adaptation between filters remains. In particular, the gradients for the subset of parameters trained on a device are calculated without consideration of the error of the remaining parameters that reside on the server. These subsets are randomly changing over time and devices, reducing the co-adaptation of this distributed training process. Add to these the fact that the data is also distributed over devices, so applying such a random scheme significantly decreases the chance that a subset of parameters is being trained together over a sizable proportion of the data.    

To further study the effects of co-adaptation on the reachable accuracy and the differences between \ac{FD} and FedRolex, we run the following experiment, using CIFAR10 with ResNet20 and $s_{\text{FD/FedRolex}}=0.25$:
\begin{itemize} [noitemsep, nolistsep]
    \item We modify \ac{FD} s.t. all devices train the same random subset per round, i.e., the same indices~$I^{(r,c)} = I^{(r)}$ per round (index $k$ is omitted for simplicity).
    \item We limit the randomness, where at each round, we arbitrarily select~$I^{(r)}$ out of a set~$\mathcal{I} = \{I_1,\ldots,I_{|\mathcal{I}|}\}$ of randomly initialized subsets that are generated once prior to training.
\end{itemize}
\begin{figure}[h]
    \centering
    \includegraphics[page=1]{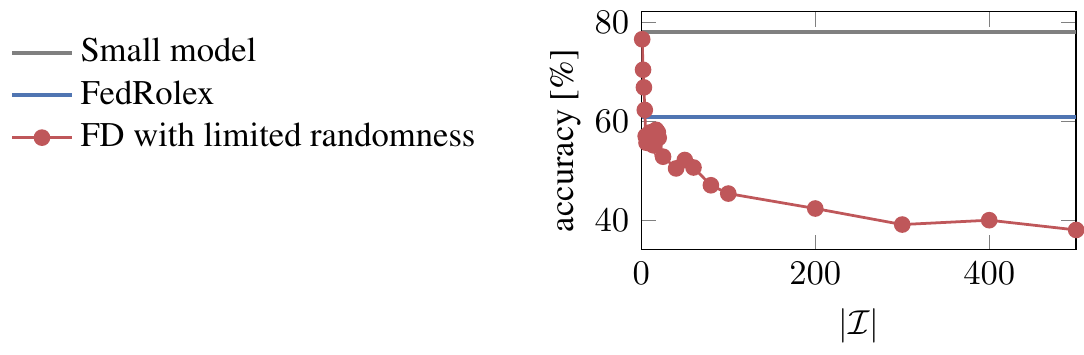} 
    \caption{\ac{FD} with limited randomness using CIFAR10 with ResNet20 and $s_{\text{FD/FedRolex}}=0.25$.}
    \label{fig:limited_randomness}
\end{figure}
We vary the randomness by varying the number of sampled subsets in~$\mathcal I$, s.t. $|\mathcal{I}| \in [0, 500]$. Thereby, the probability of a specific subset being selected is~$p = \frac{1}{|\mathcal{I}|}$. Recall that all the devices train the same submodel in a round, and there are only~$|\mathcal{I}|$ submodels that would be trained by devices over the training period. Evaluation is done with the full server model. Results are shown in~\cref{fig:limited_randomness}.

We observe the following effects: 1) The final accuracy drops proportionally to~$\sim p$. 2) In the case of~$|\mathcal{I}| = 1$, FD behaves similarly to the small model baseline, as always the same subset is used for training. We also observe that remaining untrained filters have a minor effect on the accuracy when compared with a small model. However, because of these untrained parameters, the model fails to reach higher accuracies as with \ac{SLT} (see \cref{sec:experiments}). 3) The accuracy drops with introducing more randomness to the training process (i.e., increasing~$|\mathcal{I}|$). This is as co-adaptation among parameters of the model reduces as we increase the randomness. 4) The rolling window approach of FedRolex is a special case of \ac{FD} with limited randomness 
(i.e.,~$|\mathcal{I}| = 5$ in this experiment).

\section{Ablation study maximizing~$s$ over~$F_T$}
\label{appdx:ablation_study1}
\begin{figure}
    \centering
    \includegraphics[page=1]{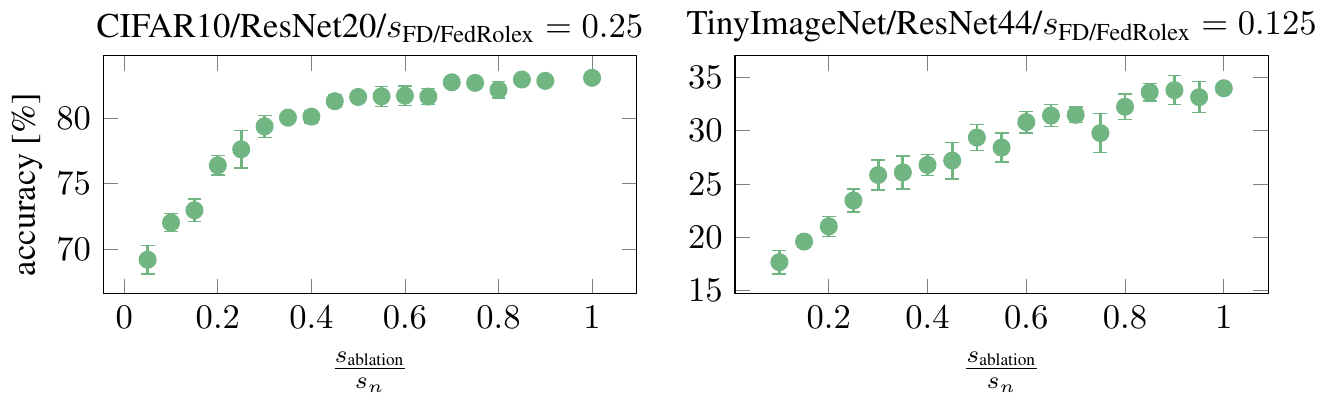} 
    \caption{Trade-off between maximizing~$s$ over $F_T$. The results show that maximizing~$s$ ($\frac{s_{\text{ablation}}}{s_n}=1$) gives the highest accuracy.}
    \label{fig:ablation_study}
\end{figure}

To justify our design choice in~\cref{sec:methodology} to maximize~$s_n$ for all steps~$n$, we conduct an ablation study, where we study the best trade-off between $s$ and $F_T$. In particular, instead of maximizing $s$, we only use fractions of the maximized~$s_n$ labeled~$s_{\text{ablation}}$. We evaluate different values for~$s_{\text{ablation}}$, i.e.~$\frac{s_{\text{ablation}}}{s_n} \in (0, 1]$. When only a fraction of the maximized~$s_n$ is used in a step, the remaining memory can be used to increase the size of $F_T$. We conduct with CIFAR10/ResNet20 and TinyImageNet/ResNet44, where all the hyper-parameters are kept the same as in~\cref{subsec:hyperparameters} (except the changes in $s_n$ and $F_T$). The final accuracy of \ac{SLT} is displayed in~\cref{fig:ablation_study}. For each run, we depict the average accuracy and standard deviation for three seeds. The results show that by maximizing~$s$ in favor of~$F_T$, \ac{SLT} reaches a higher final accuracy.

\section{Mapping of steps $N$ to rounds $R$}
\label{appdx:mapping}

\begin{figure}
    \centering
    \includegraphics[page=1]{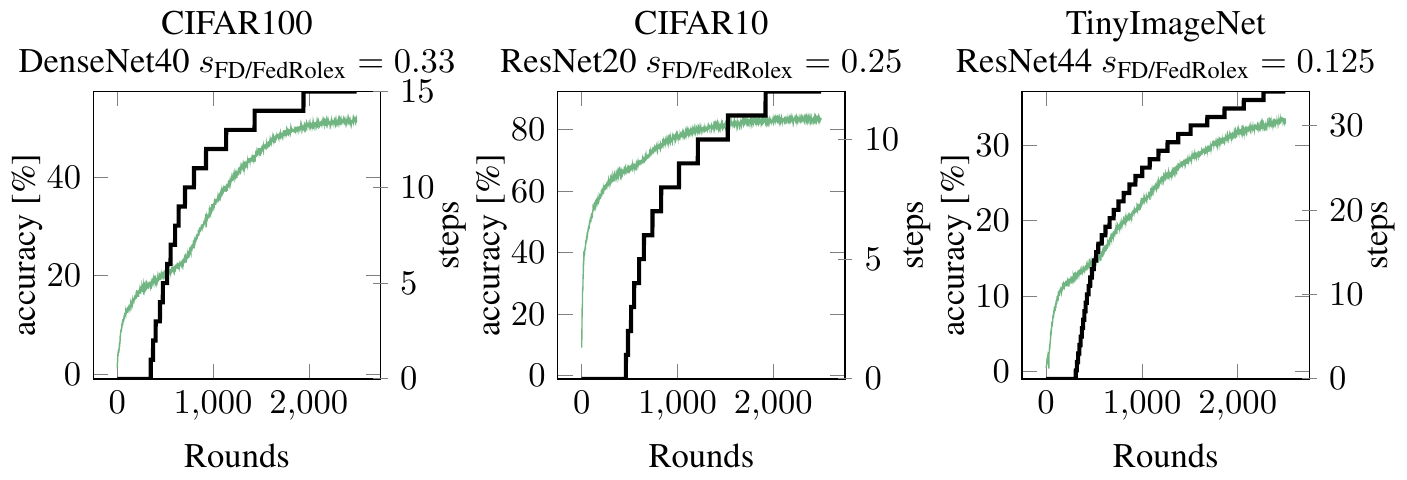} 
    \caption{Accuracy over rounds (green) and steps over rounds (black) in \ac{SLT} for CIFAR100/DenseNet40, CIFAR10/ResNet20, and TinyImageNet with ResNet44.}
    \label{fig:mapping1}
\end{figure}

Generally, any layer that is trained in \ac{SLT} should receive a sufficient amount of training to extract useful features for downstream layers, but at the same time, it should not overfit in the current configuration. The mapping of rounds~$R$ to steps~$N$ in \ac{SLT} is done proportionally to the amount of added (previously untrained) parameters to the training. Since~$N$ depends on how many steps are required until~$s=1$, the number of steps depends on the constraint level. In \cref{tab:steps}, we list~$N$ for all experiments and constraints. 
~\cref{fig:mapping1} visualizes this mapping for DenseNet40, ResNet20, and ResNet44, where \ac{SLT}'s accuracy over rounds is displayed in green while steps over rounds are displayed in black. This mapping scheme has key advantages over other techniques. Most importantly, it depends only on the \ac{NN} structure and not on the data available on the devices. Hence, it can be calculated offline prior to the training. 

\begin{figure}
    \centering
    \includegraphics[page=1]{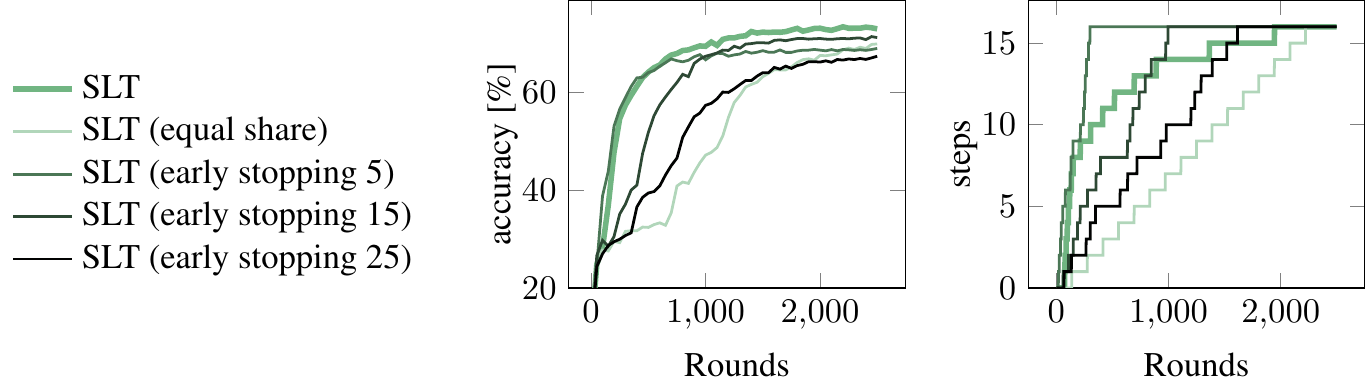} 
    \caption{Different strategies for mapping $N$ to $R$ using CIFAR10 with ResNet20 and~$s_{\text{FD/FedRolex}}=0.25$. Accuracy over rounds is displayed on the left. Steps over rounds are displayed on the right.}
    \label{fig:mapping2}
\end{figure}

We compare our mapping scheme with two other mapping schemes, one offline and one online:
\begin{itemize} [noitemsep, nolistsep]
    \item \textbf{Equal distribution:} In this scheme, we equally distribute the rounds to the steps, i.e.,~$R_n =\frac{R}{N + 1}$.
    \item \textbf{Early stopping:} In this scheme, we decide online, based on the test accuracy, when to switch. If the test accuracy on the server for a number of \ac{FL} rounds does not improve, the mapping switches to the next configuration. The number of rounds is usually referred to as \emph{patience}. We evaluate with patience $5$, $15$, and $25$.
\end{itemize}
To compare the mapping schemes, we run experiments with CIFAR10 and ResNet20, where, except for the mapping scheme, all hyperparameters are kept the same (as presented in \cref{sec:experiments}). We can observe from the results in~\cref{fig:mapping2} that our mapping scheme outperforms the others with respect to the final accuracy and convergence. Even though the early-stopping-based technique with patience $5$ increases~$n$ more aggressively, it does not result in faster convergence or higher final accuracies.

\begin{table*}
\centering
\caption{Steps $N$ in \ac{SLT} for different constraints $s$ and \ac{NN} models.}
\label{tab:steps}
\color{black}
\small{
\begin{tabular}{l c c c}
\toprule
Constraint $s$ & ResNet20 & ResNet44 & DenseNet40 \\
\midrule
0.66 & - & - & 10\\
0.5 & 8 & 14 & -\\
0.33 & - & - & 15\\
0.25 & 14 & 26 & -\\
0.125 & 16 & 14 & -\\
\bottomrule
\end{tabular}
}
\end{table*}

\section{Miscellaneous Experiments}
\label{appdx:other_results}
To evaluate how \ac{SLT} and baselines perform with more complex datasets and deeper \acp{NN}, we evaluate it with the full ImageNet~\cite{deng2009imagenet} dataset ($1.28$M images, $1$K classes). However, we downscale the images to~$3\times64\times64$ pixels to reduce the complexity of the evaluation. To account for the larger dataset, we increase the number of devices to~$|\mathcal C| = 500$ and the number of rounds to $R=4000$. All remaining hyperparameters are kept the same (\cref{sec:experiments}). We use ResNet56 to account for the more complex dataset. Results for \ac{IID} and non-\ac{IID} data are provided in \cref{tab:acc_imagenet}. We observe that the general trend of \cref{tab:acc,tab:acc_noniid} remains the same: \ac{SLT} outperforms the state of the art and the small model baseline with large margins. 

\begin{table*}
\centering
\caption{Results for \ac{IID} and non-\ac{IID} experiments with ResNet56 using Imagenet (64 $\times$ 64) are given. Accuracy in~$\%$ after $4000$ rounds of training is given.}
\label{tab:acc_imagenet}
\color{black}
\begin{adjustbox}{width=\columnwidth,center}
\small{
\begin{tabular}{l c c c c c c c c}
\toprule
Setting &
\multicolumn{4}{c}{\textbf{ResNet56/ImageNet/\ac{IID}}}&\multicolumn{4}{c}{\textbf{ResNet56/ImageNet/non-\ac{IID}}}\\ 

\cmidrule(l{1pt}r{5pt}){2-5} \cmidrule(l{1pt}r{5pt}){6-9}

$s_{\text{FD/FedRolex}}$ & 0.125 & 0.25 & 0.5 & 1.0 &
0.125 & 0.25 & 0.5 & 1.0\\
\midrule

\ac{SLT} (ours) &\textbf{24.2}$\pm$\bftab{0.2} &  \bftab{31.2}$\pm$\bftab{0.3}& 
\bftab{34.6}$\pm$\bftab{0.1} & 
\multirow{ 4}{*}{41.6$\pm$0.5} & 

\bftab{21.7}$\pm$\bftab{0.9}& 
\bftab{29.7}$\pm$\bftab{0.3} & 
\bftab{31.8}$\pm$\bftab{0.4} &
\multirow{ 4}{*}{38.7$\pm$0.3}\\

Small model&
\phantom{0}8.9$\pm$0.2 &
18.4$\pm$0.0 &
30.3$\pm$0.1 & &

\phantom{0}8.4$\pm$0.3 &
16.2$\pm$0.2 &
27.3$\pm$0.3 &\\

FedRolex~\cite{alam2022fedrolex} &
\phantom{0}3.4$\pm$0.2 &
11.4$\pm$0.3 &
21.3$\pm$0.5 & &

\phantom{0}2.8$\pm$1.1 &
10.2$\pm$0.9 &
18.4$\pm$0.4 &\\

\ac{FD}~\cite{caldas2018expanding} &
\phantom{0}0.3$\pm$0.2 &
\phantom{0}6.2$\pm$0.2 &
16.8$\pm$0.5 &  &

\phantom{0}0.1$\pm$0.0 &
\phantom{0}0.1$\pm$0.0 &
15.7$\pm$0.4 &\\

\bottomrule
\end{tabular}
}
\end{adjustbox}
\end{table*}

\section{Training Memory Measurements in PyTorch}
\label{appdx:pytorch}
We measure the maximum memory requirement for the evaluated \ac{NN} models ResNet and DenseNet using PyTorch 1.10. Specifically, we measure the size of the activations, gradients, and weights. These memory measurements are done offline (prior to training) and do not require any data.
\begin{itemize}[noitemsep, nolistsep]
    \item \textbf{Measurement of weights}: To measure the size of the weights, we sum up all tensors that are present in the \ac{NN}'s \texttt{state\_dict}.
    \item \textbf{Measurement of activations and gradients}: To measure the size of the activations that have to be kept in memory, as well as the gradients, we apply \texttt{backward\_hooks} to all relevant PyTorch modules in an \ac{NN}. Specifically, we add these hooks to \texttt{Conv2d}, \texttt{BatchNorm2d}, \texttt{ReLU},  \texttt{Linear}, and \texttt{Add} operations. If a hook attached to a module is called, we add the respective size of the activation map and the size of the calculated gradient to a global variable to add up all required activations and gradients.
\end{itemize}

\end{document}